\documentclass[number]{autarticle}
\usepackage[utf8]{inputenc}
\usepackage{amsmath}
\usepackage{amsfonts}
\usepackage{amssymb}
\usepackage{graphicx}
\usepackage{caption , subcaption}
\usepackage{lipsum}

\usepackage{float}
\usepackage{makecell}
\usepackage{color}

\begin{document}
		\begin{frontmatter}
			
			% Paper title
			\title{Clustering of Urban Traffic Patterns by K-Means and Dynamic Time Warping: Case Study\thanks{Supported by Snapp Company}}
			
			% Authors names
			\author[add1]{Sadegh Etemad}{*} 
			\ead{sadegh.etemad@snapp.cab, raziyeh.mosayebi@snapp.cab, tadeh.alexani@snapp.cab, elahe.dastan@snapp.cab, amir.salaritelmadarreh@snapp.cab, mohammadreza.jafari@snapp.cab, sepehr.rafiee@snapp.cab}
			\author[add1]{Raziyeh Mosayebi}{}
			\author[add1]{Tadeh Alexani Khodavirdian}{}
			\author[add1]{Elahe Dastan}{}
			\author[add1]{Amir Salari Telmadarreh}{}
			\author[add1]{Mohammadreza Jafari}{}
			\author[add1]{Sepehr Rafiei}{}

			\address[add1]{Map Data Chapter, Snapp Map Team, Snapp Company, Tehran, Iran}
			
			\begin{abstract}
					Clustering of urban traffic patterns is an essential task in many different areas of traffic management and planning. In this paper, two significant applications in the clustering of urban traffic patterns are described. The first application estimates the missing speed values using the speed of road segments with similar traffic patterns to colorify map tiles. The second one is the estimation of essential road segments for generating addresses for a local point on the map, using the similarity patterns of different road segments. The speed time series extracts the traffic pattern in different road segments. In this paper, we proposed the time series clustering algorithm based on K-Means and Dynamic Time Warping. The case study of our proposed algorithm is based on the Snapp application’s driver speed time series data. The results of the two applications illustrate that the proposed method can extract similar urban traffic patterns.
			\end{abstract}

		
%			\begin{AMS}
%				16W25;46L57;47B47
%			\end{AMS}

		\end{frontmatter}

	\section{Introduction}
	%\lipsum[1-3]\cite{inpr1}
    \indent Ride-sharing systems have become increasingly important and popular in recent years. They have transformed how people travel, disrupted the traditional taxi and transportation industry, and played an inevitable role in daily life. \\
    \indent Ride-sharing and map services are closely related, as map services provide the underlying infrastructure that ride-sharing systems use to identify pickup and drop-off locations, calculate distances and travel times, and optimize routing. \\
    \indent Map services provide location-based services to passengers, allowing them to quickly identify their location and find nearby drivers. This is achieved through geocoding and reverse geocoding, which allow ride-sharing systems to translate street addresses into GPS coordinates and vice versa. \\
    \indent Snapp~\cite{snapp}, one of the largest ride-sharing companies in the Middle-east, uses its map services, developed based on Open Street Map~\cite{osm}, to provide drivers with real-time traffic updates, directions, and routing information. When a passenger requests a ride, Snapp’s app uses the passenger’s location data and geocoding and reverse geocoding to find the nearest available driver. The driver then uses the map service to navigate to the passenger’s location, pick them up, and then navigate to the passenger’s destination.
    Therefore having accurate geocoding, reverse geocoding, and traffic tile is crucial to improve the user experience. Traffic tiles display the traffic flow on the roads. It shows real-time traffic, representing the traffic situation at the time of the request. Also, reverse geocoding service must be designed to provide a seamless and intuitive experience for users, with clear and accurate information presented in an easily digestible format. It is helpful to both drivers and passengers. \\
    \indent The accuracy of both reverse geocoding and traffic tile services depends on the quality and availability of data. As more data becomes available and new techniques are developed for analyzing and interpreting this data, the accuracy of these services is likely to improve. Obtaining and maintaining this data can be challenging, especially in areas where data is limited or needs to be well-maintained. These problems are the application we are going to describe in this paper. \\
    \indent In this paper, we proposed a clustering algorithm based on K-Means and Dynamic Time Warping(DTW) to improve tile colorifying and reverse geocoding. In the following, we will review some related papers about the application of clustering in urban traffic areas. \\
    \indent The traffic pattern clustering approach has been considered in various applications. For example, these algorithms can help detect anomalies in traffic patterns, such as accidents. Shafabakhsh et al.~\cite{shafabakhsh} proposed a clustering algorithm to detect the accident in the traffic pattern in Mashhad city. They used 9,331 point features for inner-city traffic accidents during one year. Eckley et al.~\cite{eckley} used various clustering techniques, including K-Means clustering, hierarchical clustering, and DBSCAN, to analyze the spatiotemporal patterns of the incidents. Their results show that the clustering methods can effectively identify spatiotemporal clusters of traffic incidents. In traffic management systems, Saeedmanesh et al.~\cite{saeedmanesh} proposed a dynamic clustering approach to analyze congestion propagation in heterogeneously congested urban traffic networks. Their proposed clustering approach obtains a feasible set of connected homogeneous components in the network called snakes, which represent a sequence of connected links with a similar level of congestion. Anbaroglu et al.~\cite{anbaroglu} proposed a Non-Recurrent Congestion (NRC) event detection method that used 424 London urban road network links. \\
    \indent Recently, there have been many studies on traffic patterns and flow estimation using different machine learning and data analytics methods. For example, one study \cite{yang} proposes a deep learning method that predicts how traffic speed propagates across the network based on the dependency and dynamics of the traffic flow. The method integrates temporal-spatial flow dependency, traffic flow dynamics, and deep learning techniques. They have achieved more than $85\%$ accuracy in predicting traffic speed changes.
    Another study \cite{leo} develops a traffic state transition matrix using the traffic data to identify urban areas with high congestion levels. Their method simplifies the clustering process and produces more interpretable results. They have shown more than $90\%$ accuracy in estimating traffic states in different urban areas which can help predict highly congested traffic flows.
    In \cite{cheng}, Cheng and et. al. applied a machine learning method to classify urban traffic patterns that can help traffic managers and travelers get more information on traffic conditions and avoid congestion. Using the ample degree of road network with other traffic data such as traffic flow, speed, and occupancy, they have shown that their proposed method, based on Fuzzy c-mean (FCM) clustering, outperforms the other approaches like support vector machines, K-Nearest Neighbor and traditional FCM clustering approach.
    In \cite{tang}, the authors extract travel patterns from large-scale vehicle trajectories. They employed the traditional DBSCAN (Density-Based Spatial Clustering of Application with Noise) algorithm with different traffic features such as spatial, directional, and temporal features. They have also proposed a statistical feature-based approach to enhance the parameter optimization process in the clustering method by fusion of the three spatial, temporal, and directional features of traffic data.\\
    \indent In \cite{ryu}, Ryu et al. proposed a clustering based traffic flow prediction method which considers the dynamic nature of spatiotemporal correlation. They have shown that the proposed method achieved a good prediction accuracy by distinguishing the heterogeneity of spatiotemporal correlations among the traffic flow. Chiabaut et al. in \cite{chiabaut} proposed a method for estimation of traffic conditions and travel time in highways. They have used Gaussian Mixture Model and K-Means algorithms for clustering of their data. Their method is tested using ten months of data collected on a French freeway and shows good results. Lin et al. in \cite{lin} suggested a technique for enhancing traffic prediction precision by screening spatial time-delayed traffic series utilizing the maximal information coefficient. Implementing their approach leads to a reduction of $23.448\%$ in Root Mean Square Error(MSE) and $14.726\%$ in Mean Absolute Percentage Error(MAPE) for the forecasted outcomes. In \cite{petrovic}, Petrovic et al. presented a hybrid soft computing model composed of two Gaussian conditional random field (GCRF) models to predict traffic speed. To evaluate the effectiveness of their proposed model, it was tested on two extensive real-world networks in Serbia. The results showed that their proposed model outperforms in terms of prediction performance.
    
    In this paper, we evaluated our proposed clustering algorithm using Snapp drivers' data in Tehran city. \\
    \indent The rest of the paper is organized as follows. In section 2, first, we describe background knowledge and then define our proposed method. Our dataset summary and system configurations are described in section 3, alongside our two applications, including tile colorification and identification of important road segments. Finally, in section 4, we describe our conclusion and future works.
    
\section{Background Knowledge and Proposed Method}
	%\lipsum[4-5]\cite{arti1}
\indent In this section, we will discuss the background knowledge and define our proposed method. 

\subsection{Background Knowledge}
    \indent Clustering is the task of dividing the data points into several groups such that data points in the same groups are more similar and dissimilar to those in other groups~\cite{estivill}. K-means is among the most straightforward unsupervised machine learning algorithms among different clustering methods~\cite{jin}. The purpose of K-means is naive: group similar data points together and try to determine the underlying patterns. To attain this objective, K-means follows a dataset's fixed number $k$ of clusters to attain this objective. The K-means algorithm has a time complexity of $O(NTK)$. $N$ represents the total number of data sets, $K$ represents the total number of partitions, and $T$ represents the number of iterations performed in the clustering process ~\cite{aristidisLikas}. The K-Means algorithm is summarized as followings~\cite{steinley}:
    \begin{enumerate}
    \item Choose the number of clusters $K$ and obtain the data points $(X_{1},X_{2},...,X_{n})$.
    \item Place the centroids $C_{1},C_{2},...,C_{k}$ randomly 
    \item Repeat steps 4 and 5 until convergence or until the end of a fixed number of iterations $(T)$
    \item for each data point, $X_{i}$:
    \begin{enumerate}
     \item find the nearest centroid $(C_{1}, C_{2}, ..., C_{k})$ 
     \item assign the point to that cluster
    \end{enumerate} 
    \item for each cluster, $j = 1, …, k$
    	\begin{enumerate}
    		\item new centroid = mean of all points assigned to that cluster
    	\end{enumerate} 
    \item End.
    \end{enumerate}

    \indent Clustering different time series into similar groups is laborious because each data point is an ordered sequence. Intuitively, the distance measures used in standard clustering algorithms, such as Euclidean distance, are often inappropriate for time series. Euclidean distance will not perform well in a time series dataset because it's invariant to time shifts, ignoring the time dimension of the data. Consider two highly correlated series, but even a one-time step shifts one. Euclidean distance would measure them as further apart~\cite{wang}. A more appropriate method is to replace the default distance measure with a metric compatible with time series, such as DTW~\cite{niennattrakul}. The algorithm behind Dynamic Time Warping is explained below with a pseudo-code~\cite{senin}:
    \begin{enumerate}
    \item We have $V_{1}=(a_{1},a_{2},…,a_{n})$ and $V_2=(b_{1},b_{2},…,b_{m})$ where $V_{1}$ and $V_{2}$ are the time series with $n$ and $m$ time points.
    \item Consider DTW a two-dimensional matrix that holds the similarity measures with $N \times M$ size.
    \item Initialize the DTW matrics as follow:
    	\begin{enumerate}
    		\item DTW(0,0) = 0.
    		\item Set each index in the first row and column to infinity.
    	\end{enumerate}
    \item Now use dynamic programming to fill each empty index in DTW as below (where d is a function to calculate distance between 2-time points):
    	\begin{enumerate}
    		\item For each $i = 2, … , n$
    		   \setlength\itemindent{15pt} \item For each $j = 2, … , m$ 
    				\setlength\itemindent{25pt}	\item dtw(i,j) = $d(V_{1}(i), V_{2}(j)) + min(dtw(i-1,j-1), 
                                 dtw(i,j-1), \\        
                                 dtw(i-1,j)) $  
         \end{enumerate}
    \item $dtw(n,m)$ is your final distance between $V_{1}$ and $V_2$. End. 
    \end{enumerate}

    \indent Computing DTW on two time series has a time complexity of $O(NM)$, where $N$ and $M$ correspond to the lengths of the two sequences being compared~\cite{senin}.

    \indent In this paper, urban traffic patterns are clustered using each road's time series of speed as the feature and the K-Means clustering method is combined with DTW as the methodology. In the next section, this method is explained, and an example of data is described.
    
    \subsection{Proposed Method}
\begin{figure}[t!]
\includegraphics[width=\textwidth]{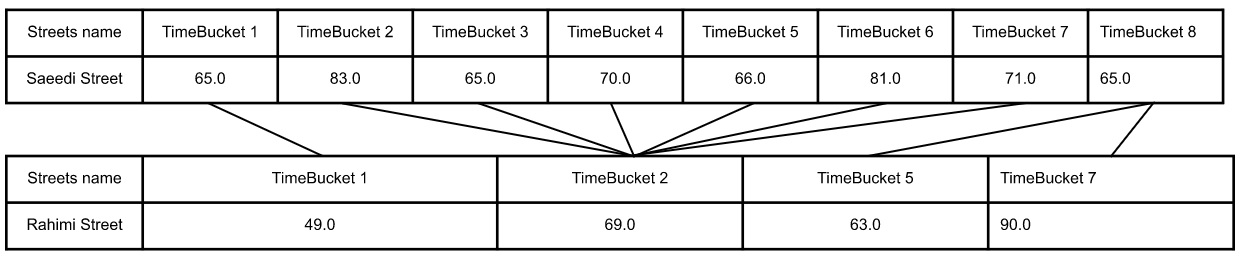}
\caption{How DTW matches 2 arrays of speed data and calculates the distance metric. Distance = 79.0
.} \label{fig2}
\end{figure}

    \indent As our world gets increasingly instrumented, sensors and systems constantly emit a relentless stream of time series data. Old approaches, such as Euclidean distance, cannot compare two data points in a time series data. Dynamic Time Warping is used to measure the similarity or calculate the distance between two arrays or time series with different length~\cite{berndt}. The idea of comparing arrays of different lengths is to make one-to-many and many-to-one matches to minimize the total distance between the two~\cite{senin}. \\
    \indent Consider we have the arrays of speed data from the Snapp taxi drivers for two different streets in Tehran. For example, Table~\ref{table1} illustrates the average speed values for Saeedi and Rahimi streets in Jordan Neighborhood. Each street has eight different time buckets that hold the average speed of the Snapp taxi drivers corresponding to that time bucket. To calculate the DTW between the two speed arrays, it is necessary first to delete the null values corresponding to time buckets with no reported speed value. Then, as you can see in Fig.~\ref{fig2}, DTW is employed to calculate the distance between the two speed arrays with different lengths.\\

\begin{table}[H]
\caption{The speed series data for 2 sample streets in Tehran. The speed values are in km/h.}\label{table1}
\resizebox{\textwidth}{!}{\begin{tabular}{|l|l|l|l|l|l|l|l|l|}
\hline
\textbf{Street's Name} & \textbf{\makecell{Time \\ Bucket 1}} &  \textbf{\makecell{Time \\ Bucket 2}} & \textbf{\makecell{Time \\ Bucket 3}} &  \textbf{\makecell{Time \\ Bucket 4}} &  \textbf{\makecell{Time \\ Bucket 5}} & \textbf{\makecell{Time \\ Bucket 6}} &  \textbf{\makecell{Time \\ Bucket 7}} &   \textbf{\makecell{Time \\ Bucket 8}}
\\
\hline
Saeedi Street & 65.0 &  83.0 & 65.0 & 70.0 & 66.0 & 81.0 & 71.0 & 65.0 \\
\hline
Rahimi Street & 49.0 &  69.0 & NA & NA & 63.0 & NA & 90.0 & NA \\
\hline
\end{tabular}}
\end{table}

\indent Fig.~\ref{fig2} shows the performance of DTW on the two speed arrays after dropping the null values. The two speed arrays are fed to the DTW algorithm. Fig.~\ref{fig2} shows how DTW matches the arrays together. The distance between the two speed arrays is calculated and used instead of euclidean distance in K-Means clustering.

\indent As described in the previous two sections, DTW is a suitable metric for K-Means clustering for extracting similar patterns between the speed time series data in our case study. It will remove the necessity of imputing the missing values, which induces a calculation error to the data and consequently affects the clustering results.

\section{Evaluation and Experimental Results}
\indent This section presents the experimental results of our proposed method for two applications: "Street Tile Colorification" and "Identification of Important Road Segments." The details of the dataset are provided in sub-section 3-1. Furthermore, we discuss applying our method to these applications in sub-sections 3-2 and 3-3.

\subsection{Dataset Description and Configuration}
\indent The "Dataset Description and Configuration" section provides information about the dataset used in this study, including its summary, preprocessing details, and system and framework configuration. The sub-sections 3-1-1, 3-1-2, and 3-1-3 provide an overview of the dataset, its preparation for analysis, and the hardware and software requirements.

\subsubsection{Dataset Summary}
\indent The dataset used in this study consists of records of street-level vehicle speed and timestamp information collected from a sample of 2.4 million unique time series data for two weeks. The dataset was collected from various residential, commercial, and highway streets. Data were collected from each vehicle's driver using GPS-enabled devices like mobile phones and then aggregating speeds for each street based on the median statistical measure. Table~\ref{table2} describes the features included in the dataset.

\begin{table}[H]
\caption{The features of our Snapp drivers' dataset.}\label{table2}
\resizebox{\textwidth}{!}{\begin{tabular}{|c|c|}
\hline
\textbf{Feature Name} & \textbf{Feature Explanation}\\
\hline
Street ID &  A unique identifier for each street on which the speed measurement was recorded.\\
\hline
Speed (km/h) &  The aggregated vehicle speeds in kilometers per hour. \\
\hline
Time Bucket &  \makecell{The time during which the speed measurement was recorded. \\ The time bucket is divided into 15-minute intervals, starting at 00:00 and ending at 23:45.\\ The selection of a 15-minute interval enabled us to capture significant variations in speed \\ while preserving a manageable level of noise in the data, thus maintaining a balanced approach.}
  \\
\hline
\end{tabular}}
\end{table}

%\begin{tabular}{@{}c@{}} The time during which the speed measurement was recorded. \\ The time bucket is divided into 15-minute intervals, starting at 00:00 and ending at 23:45.

\indent The data were collected continuously during the two weeks, each row representing a unique sample of a street’s aggregated speed, street id, and corresponding time bucket.

\subsubsection{Data Pre-processing}
\indent Before conducting the analysis, we performed several pre-processing steps on the dataset. These steps included data cleaning, where we removed any outliers or duplicated data and corrected any errors in the dataset. We used Spark's ~\cite{spark} built-in data cleaning functions to remove any missing values and duplicated records. The minimum and maximum speed thresholds are used to remove the speed values out of these ranges. We also retained streets with more than two speeds in their weekly time buckets to account for data scarcity, given that the data is derived from Snapp drivers and the streets they route.
After cleaning the data, we performed several transformation steps, such as filtering, grouping, and joining the data using Spark SQL. For example, we manually generated a weekly time bucket to view trends within specific weeks, days, or hours. This feature allowed us to cluster streets based on their traffic patterns during specific time slots. Each week has 672 buckets of 15 minutes, and the final dataset contains 673 columns and 2.4M samples after adding the weekly time bucket feature and aggregating the data. \\
\indent To estimate the speed in every 672-time bucket, we used the GPS location of Snapp drivers who passed the street in that time bucket (note that not every road has a driver in every time bucket). We then defined a new Filling-rate feature, the ratio of the time buckets in which at least a Snapp driver has passed, divided by 672. As a result, the significant streets on the Snapp side have a Filling-rate of more than $33.33\%$. In the rest of the paper, we removed the streets with a Filling-rate less than this value. Using this filter, we can focus on roads with higher traffic activity, including streets where Snapp drivers pass more frequently. \\
\indent The final dataset provides a comprehensive and representative sample of street-level vehicle speed data and time bucket information, making it suitable for analyzing patterns and trends in traffic patterns and their relationship with the time of the day and street characteristics. The external GIS data is joined to the final dataset using street ID. As a result, features such as the street length, type of road, and maximum and average speed are added for each street ID. \\
\indent Furthermore, in this study, we used Apache Spark, a robust extensive data processing framework, to parallelly process a large dataset. We chose to use Spark because it provided scalability, speed, flexibility, ease of use, and reproducibility in processing our large dataset. This allowed us to obtain more accurate and reliable results, which would have been difficult to achieve using traditional data processing methods. Additionally, Spark's ability to handle real-time data allowed us to analyze and make sense of our data on the fly, making insights that were impossible to detect.

\subsubsection{System and Framework Configuration}
\indent During this study, we used a virtual machine configured as follows:
\begin{itemize}
\item \textbf{Hardware}: The virtual machine has a single CPU with 12 cores and 32 GB of RAM. The storage capacity was 250 GB, and the CPU was an Intel(R) Xeon(R) Platinum 8268 processor with a clock speed of 2.9 GHz. \\

\item \textbf{Software}: The virtual machine ran on Ubuntu 20.04.5 LTS operating system, and we used PySpark version 3.3.1 for distributed processing of the large dataset. We also installed additional packages, Python 3.8, Jupyter Notebook, and tslearn, for time-series data pre-processing and visualization.
\end{itemize}

\begin{figure}[t!]
\begin{center}
\includegraphics[scale=.55]{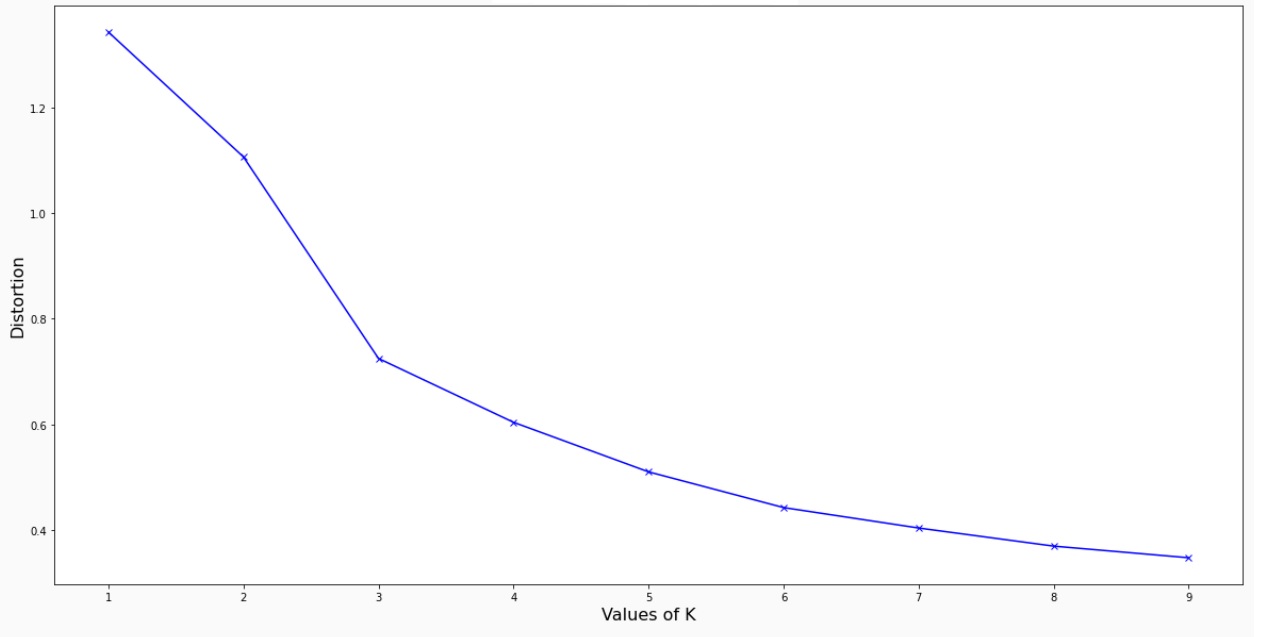}
\caption{The results of the elbow method on the dataset. Cluster number 3 is selected as the final number for clustering.} \label{fig3}
\end{center}
\end{figure}

\indent This configuration was used to process a large dataset of over 2.4 million records and allowed us to perform the data processing tasks promptly and efficiently.
For the K-Means clustering, we evaluated the performance of the clustering using different numbers of clusters (k). The elbow analysis was used to confirm the optimal number of clusters which measures how well each point is assigned to its cluster compared to other clusters ~\cite{kodinariya}. Fig.~\ref{fig3} shows the elbow method drawn to choose the best cluster number. Cluster number 3 is used as the final value for the clustering. Each of these 3 clusters represent a different kind of street. One consists of alleys and neighborhood streets with 1 or 2 lanes, speed limit of 30 km/h and residential class in OSM. Another cluster members are usually highways and freeways with 4 lans, max speed of 100 km/h and trunk road class in OSM like Hemat highway, and the last cluster includes mostly avenues and boulevards with 2 or 3 lanes, max speed of 50 km/h and secondary class in OSM like Marzdaran boulevard.

\subsection{Application 1: Street Tile Colorification}
\begin{figure}[t!]
\begin{center}
\includegraphics[scale=.55]{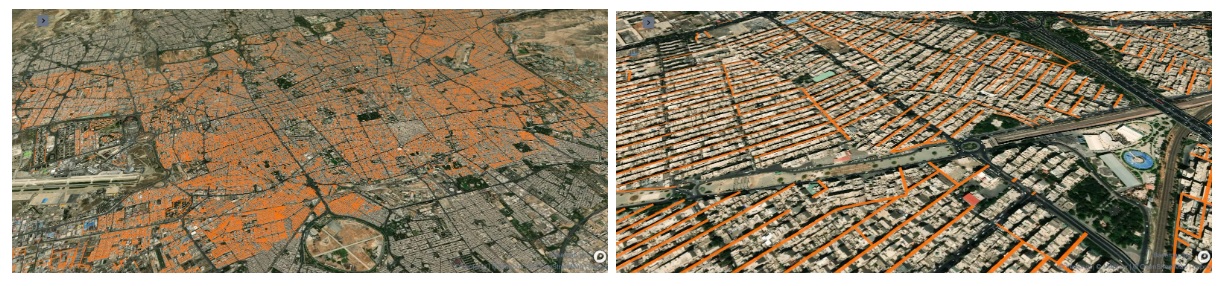}
\subcaption{}
\includegraphics[scale=.55]{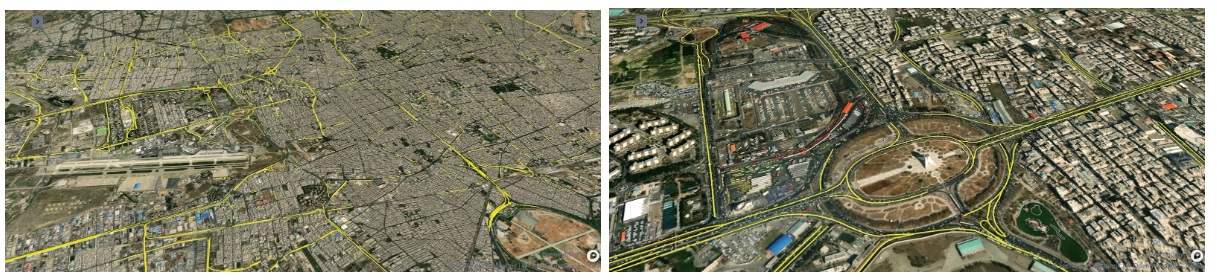}
\subcaption{}
\includegraphics[scale=.55]{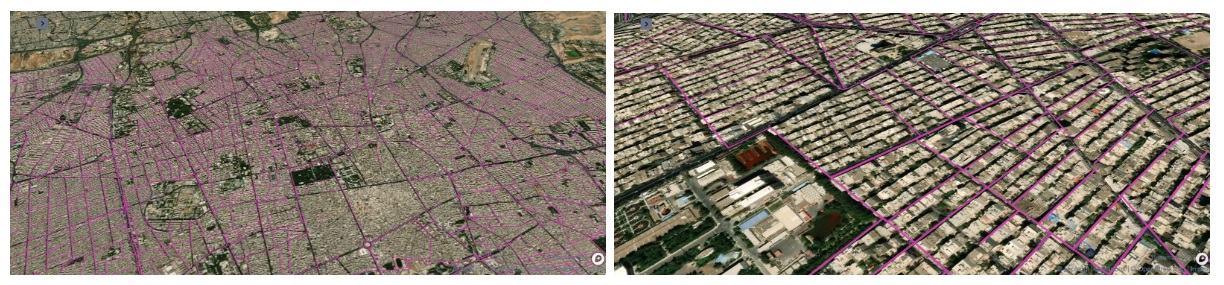}
\subcaption{}
\end{center}
\caption{\label{fig4} Visualization of Streets Clustered Based on Similar Traffic Characteristics (a) cluster 1 where majority of streets are Less Extensive (b) cluster 2 where majority of streets are Freeways and Motorways (c) cluster 3 which majority of streets are main streets.}
\end{figure}

\indent Clustering is a suitable approach to group the streets with similar traffic patterns such as traffic volume, max speed, and type of road. Among different purposes for clustering traffic patterns, we concentrated on the imputation of speed for those streets with no reported values. \\
\indent Consider two streets that have similar traffic volumes and speeds. As a result, it is reasonable to assume that these streets have similar speed values at different times of the day and week.  Consequently, if the speed of one street is known for a particular time, it can be used for imputing the missing speed values for the other street. Once the streets are grouped into clusters, the speeds can be imputed by taking the average of the speeds of all the streets in the same cluster. \\
\indent We apply our proposed method to our dataset and get the following results. Fig.~\ref{fig4} shows that the roads can be clustered into three major groups. Fig.~\ref{fig4}(a) shows the type of roads that are almost neighborhood streets or alleys, Fig.~\ref{fig4}(b) represents the main roads like highways, freeways, or main squares, and Fig.~\ref{fig4}(c) shows a third type of streets that mostly contain avenues and boulevard and have an average traffic volume.

\begin{figure}[t!]
\begin{center}
\includegraphics[scale=0.6]{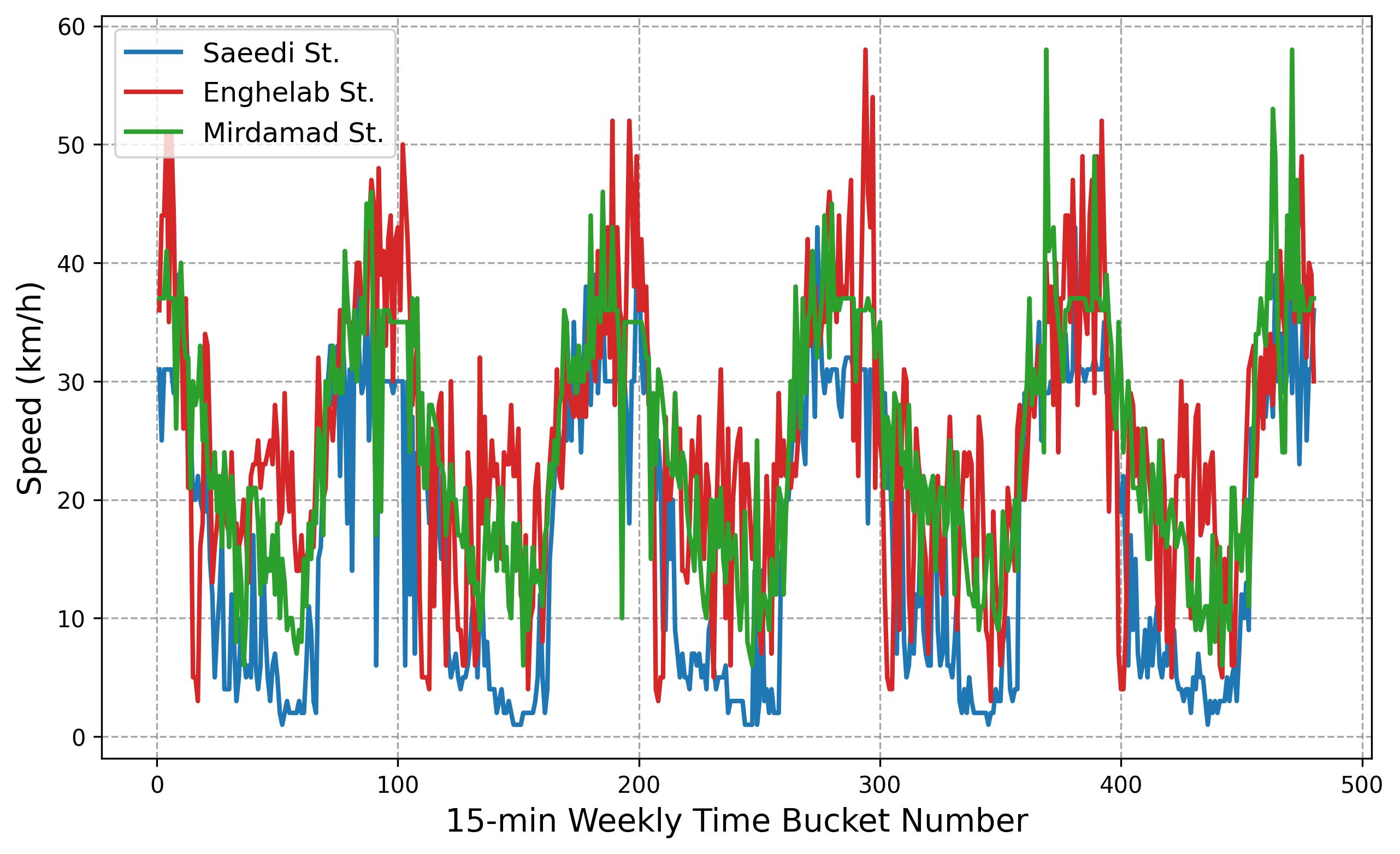}
\caption{Comparison of Weekly Traffic Speed Patterns on Primary Streets (Mir-damad and Enghelab) and a Residential Street (Saeedi) with Similar Characteristics (speeds in km/h).} \label{fig5}
\end{center}
\end{figure}

\indent This clustering categorizes roads based on their traffic behavior, so two roads with different characteristics can be in the same group. Fig.~\ref{fig5} shows the traffic behavior of three roads that are in the same cluster. Although they have different OSM road classes, their traffic pattern shows similar behavior during the working days of the week.

\indent Traffic Tile reflects the traffic congestion on the road network, presented to the user by color-coded lines drawn along the affected streets and roads. The traffic congestion displays four congestion levels:
\begin{itemize}
\item Free Flow (green): Road segments where the average speed of the traffic is around the expected free flow speed, mostly due to a small volume of cars.
\item Heavy (yellow): Road segments with a higher volume of cars, leading to an average speed that is less than the expected free flow speed.
\item Queuing (red): Road segments with a very high volume of cars leading to an average speed closer to stopping and reducing the distance between two successive vehicles, which would have the effect of creating a queue.
\item Blocked (black): Road segments that are blocked, due to temporary issues such as construction sites and accidents.
\end{itemize}

\begin{figure}[!htb]
   \begin{minipage}{0.48\textwidth}
     \centering
     \includegraphics[scale=.55]{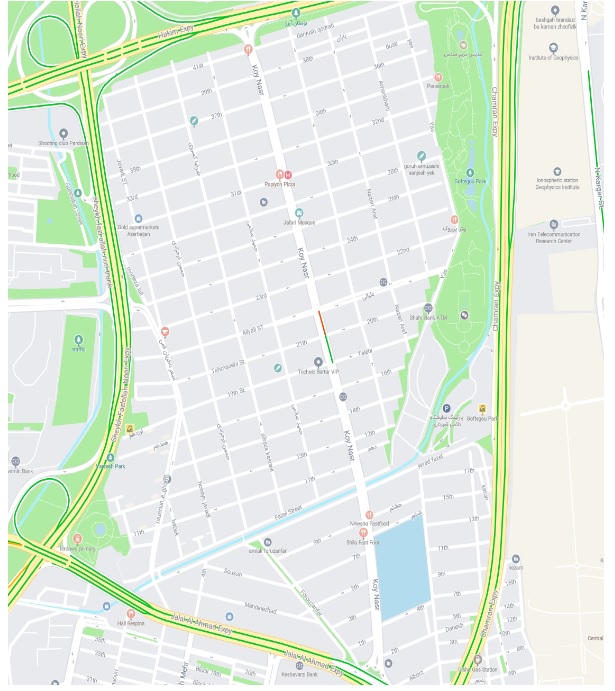}
     \subcaption{}
   \end{minipage}\hfill
   \begin{minipage}{0.48\textwidth}
     \centering
	\includegraphics[scale=.55]{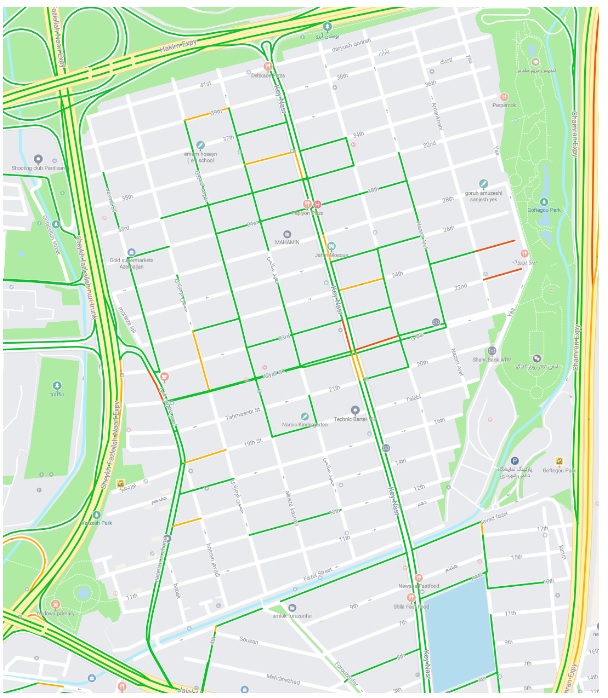}
     \subcaption{}
   \end{minipage}
\caption{\label{fig6} Snapp application's traffic tile, (a) tile without proposed method, (b) tile with proposed method.}
\end{figure}

\indent Based on the congestion levels described, users will be informed of the state of the roads. This clustering led us to add or change the color of some roads on the map for better visualization of traffic. Fig.~\ref{fig6} shows we have been able to colorify some roads in tile, which we weren't able to do before since few drivers pass them.

%\begin{figure}[t!]
%\begin{center}
%\includegraphics[scale=.55]{6-1.jpg}
%\subcaption{}
%\includegraphics[scale=.55]{6-2.jpg}
%\subcaption{}
%\end{center}
%\caption{\label{fig6} Snapp application's traffic tile, (a) tile without proposed method, (b) tile with proposed method.}
%\end{figure}

\subsection{Application 2: Identification of Important Road Segments}
\indent The problem of classifying road segments based on their importance is very common in geo-related fields. There is an application in reverse geocoding in which, given the geographical location of a pin location (lat, long), the full address can be derived.  For each point in the map, reverse Geo-code generates addresses as follows.
\begin{enumerate}
\item Create a road network graph for each route.
\item Traverse nodes like Breadth-first search and start generating address
\item Address will be complete if we reach a primary street.
\end{enumerate}

\indent There are different types of roads in the Open Street Map ~\cite{osm}. The typical road types in OSM are Primary, Secondary, Tertiary, and Trunk. A significant problem with reverse geocoding will be the possibility of the absence of any primary roads in the neighborhood of the location where we intend to generate an address. Thus, in such cases, the task of generating addresses will be continued until a primary road is reached. However, this primary road may be a highway that has no access to a local street. As a result, the generated address will be useless since the driver does not have direct access to the location from an accessible address. This kind of address, in reverse geo-coding, is called an inaccessible address.\\
\indent Based on our investigations, we found a few essential roads near the pin locations that are well-known and can be used for address generation; however, they are not the primary routes. In order to resolve this problem in reverse Geo-coding, we decided to identify the non-primary important roads and streets that are quite well known to the user and can be used as a suitable replacement for an inappropriate primary road. \\
\indent To identify important non-primary roads, we used the time series of speeds described in the previous section. Besides the speed time series (672 samples), filling rate, type of street ( primary, secondary, residential, …), and max speed are used in this analysis. The analysis steps are as follows:
\begin{enumerate}
\item A standard scaler is applied to the current feature vector (using 672 samples from the speed time series besides the filling rate and max speed). This scaler removes the scale difference between different features and results in faster convergence in our clustering method.
\item The primary road segments are separated from the clustering. Using the features of speed time series, max speed, and filling rate, a representative feature vector is calculated by averaging the feature vector of the primary roads. As a result, a feature vector is calculated for primary road segments.
\item The secondary road type is selected to apply the clustering method. The other road types are not considered a replacement for primary streets.
\item After preprocessing, K-means clustering is applied on secondary road segments by using the dynamic time warping method as the metric for clustering. The dynamic time-warping metric considers the features of time series data for calculating distance.
\end{enumerate}

\indent The elbow method ~\cite{kodinariya} is used in order to select the number of clusters. The elbow method shows that cluster number 3 is a suitable candidate for clustering. We examined the value of 3 and 4 for clustering. Besides the elbow method, we also checked the result of clustering, such that the feature vector of each cluster must have a distinctive pattern. As a result, the secondary road types are grouped into 3 clusters. One of these clusters has a centroid that has the least distance with the feature vector of primary road segments. The streets of that cluster are suitable candidates since they have the most similar traffic patterns with primary road segments. \\
\indent A few examples of the secondary roads in the selected cluster are shown in Table~\ref{table3}. These streets are chosen in Tehran to be clear for most readers.

\begin{table}[H]
\caption{Examples of selected roads in clustering.}\label{table3}
\begin{center}
\begin{tabular}{|l|l|l|l|}
\hline
\textbf{Road Class} & \textbf{Filling Rates$\%$} & \textbf{Name} & \textbf{County}\\
\hline
\small Secondary &  93.45238 & Vahid Dastgerdi & Tehran \\
\hline
\small Secondary &  91.815475 & North Jannat-Abbadi & Tehran \\
\hline
\small Secondary &  93.60119 & Seyed Jamallodin Asadabadi & Tehran \\
\hline
\small Secondary &  95.53571 & Mollasadra & Tehran \\
\hline
\small Secondary &  94.49405 & Marzdaran Boulevard & Tehran \\
\hline
\small Secondary &  98.66071 & Ayatollah Beheshti & Tehran \\
\hline
\end{tabular}
\end{center}
\end{table}

\indent The proposed method identifies the important secondary roads. As a result, for generating addresses, these newly identified road segments can be used instead of primary segments, creating more user-friendly addresses. To illustrate the effectiveness of the proposed clustering approach, for a pin location on the map in Tehran city, two instances of addresses are presented before and after applying the proposed method. \\

\begin{figure}[t!]
\begin{center}
\includegraphics[scale=0.6]{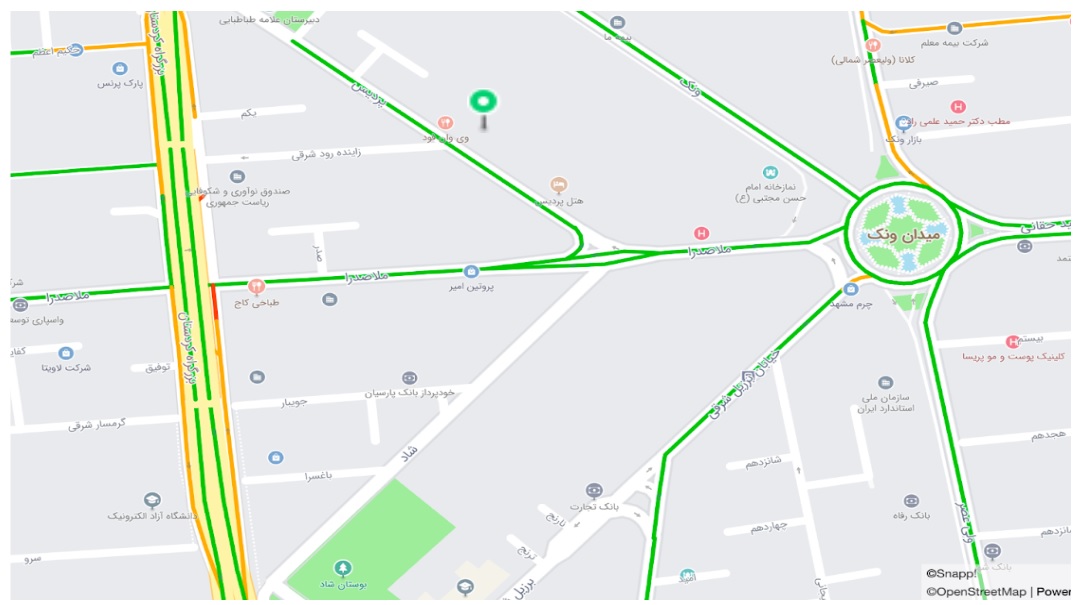}
\caption{Pin location with Latitude: 35.757339 and Longitude: 51.407911} \label{fig7}
\end{center}
\end{figure}

\indent As you can see in Fig.~\ref{fig7}, before applying the results of the clustering, the generated address in the vicinity of 35.757339, 51.407911 is: 
\begin{itemize}
\item Tehran, Vanak, Kordestan Expressway, Molla Sadra Street before Sadr, Company X.
\end{itemize}
As it is shown in Fig.~\ref{fig7}, there is currently no route from Kordestan Expressway to MollaSadra street due to the street's one-way traffic in the direction of Kordestan Expressway. This type of address confuses the passengers and drivers who use the address for navigation and route finding. However, after applying the proposed method,  the following address is generated for the exact location:
\begin{itemize}
\item Tehran, Vanak, Molla Sadra Street before Sadr, Company X.
\end{itemize}
This method shows that the new address is more accurate and closer to how a human writes an address. Kordestan Expressway is a primary street used in the address-generating approach logic. However, as there is an important secondary segment like Molla Sadra Street, it is not necessary to use Kordestan Expressway and create confusion. This is just one example of how our clustering algorithm can reveal hidden patterns in address data, making it easier to perform targeted analyses and identify areas of interest.

\indent In conclusion, the clustering of street features can identify well-known roads without the need to manually tagging each road class. It will result in better-generated addresses and speed up reverse-geocoding time.

\section*{Discussion}
\indent In this paper, the clustering of streets based on the reported speed values from Snapp drivers is explained. Two significant applications were discussed. The first one is estimating the speed for non-reported time buckets based on the clustering results, which were used to visualize traffic volume on the map based on the similarity of speed patterns between streets. The second application was the identification of important streets to use for generating shorter addresses. The K-mean clustering with DTW metric was employed to cluster the speed time series in both applications. The DTW metric calculates the similarity between the time series of speed with different lengths for different road segments. The difference between the length of the speed time series results from having no reported values in some of the time buckets. \\
\indent Another approach for clustering speed time series is to impute the missing values (the time buckets with no reported speed values) and then use an ordinary K-means clustering with the Euclidean metric. To justify the reason for not choosing this method, we must refer to the disadvantages of missing values imputation, which induces a bias to our speed values. This bias will then be propagated to our clustering results, affecting the traffic pattern. Our clustering results with K-means clustering and DTW metrics illustrate the effectiveness of the proposed method for extracting similar traffic patterns.

\section*{Conclusion and Future Works}
\indent In this paper, we discussed two significant applications for extracting similar urban traffic patterns. The time series of speed is an important feature that reflects the traffic pattern in each road segment. We employed the K-means clustering method with the DTW metric to cluster the road segments based on their speed time series and filling rate. The results show that similar traffic patterns between road segments are extracted successfully and can be used for missing value imputation and address generating in reverse geocoding. In future works, we will use cluster labels as a feature in the speed prediction system, and we will incorporate those predicted speeds into our estimation of the time of arrival (ETA).

\section*{Acknowledgment} This paper and the research behind it would not have been possible without the exceptional support of Snapp Map’s managers especially Ali Karami, Sina Bakhtiari, and Mohammad Julaiee. We would also like to thank the Snapp Company for their support.

\end{document}